\documentclass{article}
\usepackage{spconf,amsmath,graphicx}

\usepackage{enumitem}
\setlist{nosep, leftmargin=14pt}

\usepackage{mwe} % to get dummy images

\usepackage{booktabs} % for \toprule, \midrule, \bottomrule
\usepackage{comment}
\usepackage{url}

\title{Stroke Lesion Segmentation in Clinical Workflows: A Modular, Lightweight, and Deployment-Ready Tool}
\name{Yann Kerverdo$^{1}$ \quad Florent Leray$^{1}$ \quad Youwan Mah\'e$^{1,2}$ \quad 
Stéphanie Leplaideur$^{1,3,4}$ \quad Francesca Galassi$^{1}$}

\address{%
$^{1}$ Univ Rennes, Inria, CNRS, Inserm, IRISA UMR 6074, Empenn, Rennes, France \\
$^{2}$ Siemens Healthineers, Courbevoie, France \\
$^{3}$ CHU Rennes, Physical Medicine and Rehabilitation Department, Rennes, France \\
$^{4}$ Centre de Kerpape, Ploemeur, France\\
}
\begin{document}
%\ninept
%
\maketitle
\begin{abstract}
Deep learning frameworks such as nnU-Net achieve state-of-the-art performance in brain lesion segmentation but remain difficult to deploy clinically due to heavy dependencies and monolithic design. We introduce \textit{StrokeSeg}, a modular and lightweight framework that translates research-grade stroke lesion segmentation models into deployable applications. Preprocessing, inference, and postprocessing are decoupled: preprocessing relies on the Anima toolbox with BIDS-compliant outputs, and inference uses ONNX Runtime with \texttt{Float16} quantisation, reducing model size by about 50\%. \textit{StrokeSeg} provides both graphical and command-line interfaces and is distributed as Python scripts and as a standalone Windows executable. On a held-out set of 300 sub-acute and chronic stroke subjects, segmentation performance was equivalent to the original PyTorch pipeline (Dice difference $<10^{-3}$), demonstrating that high-performing research pipelines can be transformed into portable, clinically usable tools.
\end{abstract}
\section{Introduction}
\label{sec:intro}
Stroke is a leading cause of long-term disability worldwide~\cite{feigin2021global}. In the sub-acute and chronic phases, structural magnetic resonance imaging (MRI) is routinely used to assess lesion extent, monitor recovery, and guide rehabilitation strategies~\cite{bernhardt2017agreed}. Accurate lesion segmentation is essential for quantifying brain damage and linking imaging to functional outcomes. Manual delineation, however, is time-consuming and subject to inter-rater variability~\cite{ahmed2023appraisal}, motivating automated methods.
Recent deep learning advances, notably nnU-Net~\cite{isensee2021nnunet}, have achieved state-of-the-art performance in post-stroke lesion segmentation on datasets such as ATLAS~v2.0~\cite{liew2018atlas}. Di~Matteo~et~al.~\cite{dimatteo2025stroke} further trained stroke-specific nnU-Net~v2 models on T1-weighted (T1w) and T1w+FLAIR MRI, releasing pretrained weights and inference code. We use these models as a case study to present a general framework for translating nnU-Net pipelines into lightweight, deployable tools applicable across medical imaging tasks.

Existing toolkits such as TorchIO~\cite{torchio2021} and pymia~\cite{jungo2020pymia} support reproducible preprocessing for research, while system-level frameworks like MONAI~Deploy~\cite{monai_deploy} and PACS-AI~\cite{theriault2024pacsai} enable integration with DICOM and PACS infrastructures. However, these solutions do not address a key practical challenge: transforming a pretrained research model into a lightweight application that can run directly on a clinical workstation.

This work addresses this limitation by introducing \textit{StrokeSeg}~\cite{StrokeSegUserGuide2025}, a versatile application that translates high-performing research models into deployable, clinically usable tools. Specifically, it (i) restructures the nnU-Net pipeline into independent, interoperable components, (ii) improves computational efficiency and portability through ONNX Runtime and \texttt{Float16} quantisation, and (iii) delivers a reproducible implementation suitable for clinical research workflows.

\section{Methods}
\label{sec:methods}
\subsection{Architecture overview}
\begin{figure}[htb]
  \centering
  \includegraphics[width=0.8\linewidth]{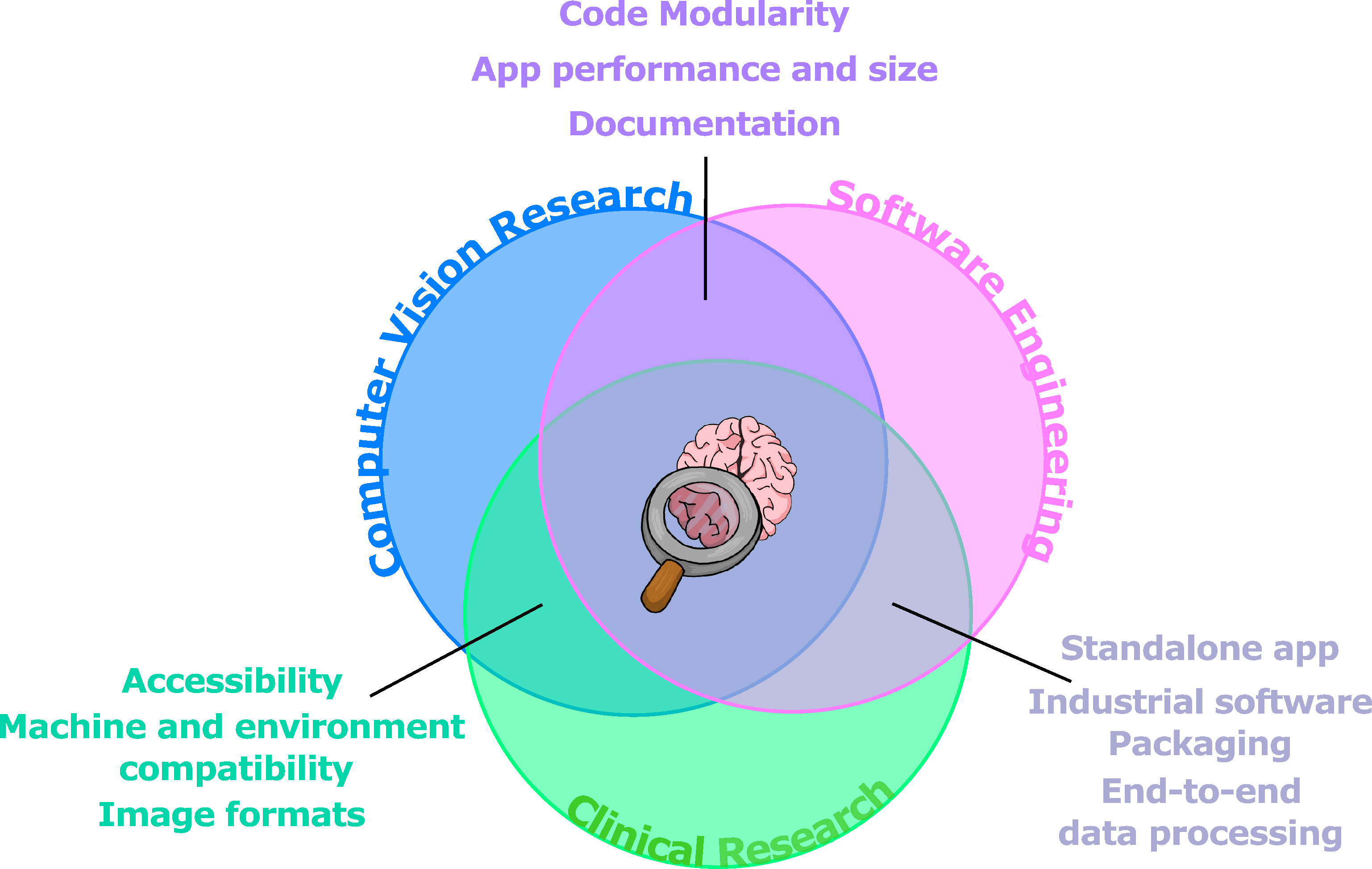}
  \caption{Challenges at the intersection of research, engineering, and clinical practice. Each overlap highlights domain-specific constraints addressed in this work.}
  \label{fig:venn}
\end{figure}

Our goal was to transform the stroke-specific nnU-Net–based segmentation pipeline of Di Matteo et al.~\cite{dimatteo2025stroke} into a lightweight, portable, and reproducible research tool. 
The design needed to satisfy two distinct user groups: (i) clinical researchers, who require a simple, installable application for routine analysis, and (ii) computer vision researchers, who expect transparency and modularity for further development. Bridging these domains - computer vision research, software engineering, and clinical practice - introduces specific technical and usability constraints (Fig.~\ref{fig:venn}). To address them, the development was structured along three axes: (1) modular architecture, separating preprocessing, inference, and postprocessing; (2) performance optimisation through runtime and model-level improvements; and (3) packaging for reproducible, cross-platform deployment. Two packaging targets were defined to reflect typical usage scenarios: Windows systems for clinical research applications and Linux systems for computer vision research and high-performance computing (HPC) environments.

\subsection{Scientific package}
The original nnU-Net–based pipeline~\cite{isensee2021nnunet, dimatteo2025stroke} was refactored into three independent stages.

\textbf{Preprocessing.} 
Brain extraction and spatial normalisation are performed with the Anima toolbox~\cite{anima_empenn}, wrapped in a dedicated module. Both monomodal (T1w) and bimodal (T1w + FLAIR) inputs are supported; FLAIR volumes, when provided, are rigidly registered to T1w. Intermediate files are organised following BIDS conventions to ensure traceability and interoperability.

\textbf{Inference.} 
Pretrained models from~\cite{dimatteo2025stroke} were exported to the ONNX graph format and executed with ONNX Runtime, removing heavy training-only dependencies. As in nnU-Net~\cite{isensee2021nnunet}, the inference module implements patch-based prediction and Gaussian-weighted merging. Inference can be executed on either CPU or CUDA backends, and only the required dependencies are packaged for each target.

\textbf{Postprocessing.} 
Network logits are converted to probability maps via softmax. The lesion channel produces a probability map and a binary mask after configurable thresholding (default~0.5). Optionally, outputs can be saved in MNI space, with inverse transforms applied to recover subject-space results.

\begin{figure*}[t]
\centering
  \includegraphics[width=.9\linewidth]{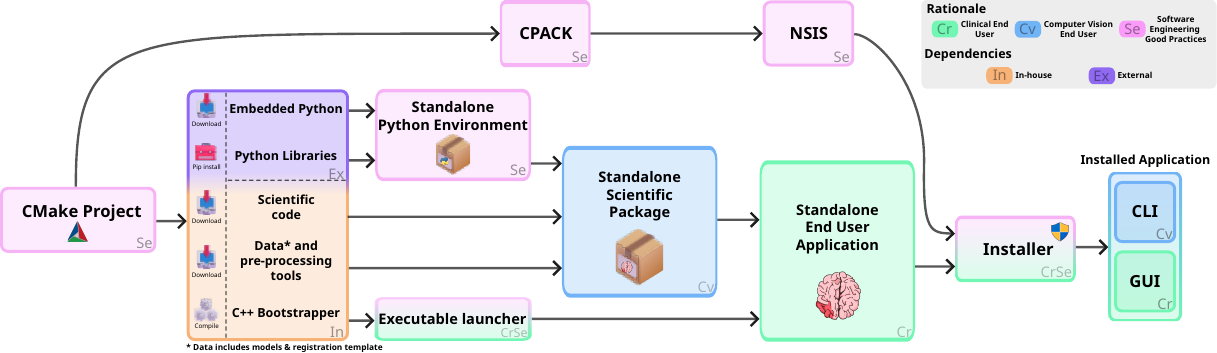}
  \caption{Overview of the \textit{StrokeSeg} packaging workflow. The CMake project compiles the C++ launcher and gathers the scientific Python modules, embedded Python environment, and pretrained models. CPack then builds a platform-specific installer (e.g., NSIS for Windows) supporting both graphical and command-line execution. The resulting package can be launched directly by end users without manual environment configuration.}
  \label{fig:framework}
\end{figure*}

\subsection{Performance-oriented inference}
Inference performance was compared between the native PyTorch implementation and the ONNX Runtime version. The ONNX-based approach markedly reduced software dependencies and overall package size (Sec.~\ref{sec:size}) while preserving accuracy (Sec.~\ref{sec:accuracy}). Model weights were converted from \texttt{Float32} to \texttt{Float16}, reducing storage requirements by about 50\%.
After preliminary optimisations, brain extraction emerged as the main runtime bottleneck. To address this, \textit{StrokeSeg} introduces two optimisations: (i) a dedicated \textit{Brain Extraction Only} mode, and (ii) a caching mechanism allowing either re-use of previously processed data or loading of externally prepared brain-extracted volumes, thereby avoiding redundant computations.
During inference, GPU acceleration is used automatically when a compatible CUDA runtime is available; otherwise, execution seamlessly falls back to the CPU. This behaviour is essential in clinical environments, where most workstations lack discrete CUDA GPUs.

Runtime measurements (Sec.~\ref{sec:size}) were obtained on two representative hardware configurations: (1) an Ultrabook laptop (Intel i7-10610U CPU, 16~GiB RAM, no dedicated GPU) representing typical clinical research hardware; and (2) a workstation laptop (Intel i7-13850HX CPU, NVIDIA RTX~5000 Ada GPU, 32~GiB RAM) representing a computer vision research setup. Ten subjects T1w were processed with the monomodal pipeline using the 16-bit quantised model executed via ONNX Runtime. Runtime measurements were aggregated across three processing stages: brain extraction, registration, and model inference.
\subsection{Packaging and distribution}
\subsubsection{Main packaging target} 
For classical Windows-based clinical environments, \textit{StrokeSeg} is packaged as a fully self-contained application using Embedded Python and an NSIS installer generated via CMake/CPack. Self-containment is essential for clinical end users, who typically have limited interaction with command-line tools. To ensure a user-friendly experience, a lightweight C++ bootstrapper launches the Python graphical interface, hides the console window, and presents the software as a single executable (Fig. \ref{fig:framework}). This bootstrapper serves as the main entry point, supporting three modes: command-line (headless), graphical (GUI), and combined debug mode. 
The installer bundles all required dependencies, removing the need for a system-wide Python installation or internet access to download models. This fully contained build ensures reproducibility, as all package versions are fixed at compile time, and avoids conflicts with any existing Python environments. The resulting package supports OS-compliant installation and uninstallation procedures, providing clean integration and removal. 
\subsubsection{Secondary packaging targets.} For Linux-based research environments, the scientific package is deployed via a Bash script that installs all dependencies in the appropriate directories within a Python virtual environment. This setup requires root privileges, internet access, and sufficient Bash expertise to manage potential edge cases.

\subsection{Software design and extensibility}

Model management features include ONNX model registration and selection. The architecture enables transparent integration of any 3D nnU-Net model after ONNX conversion, allowing rapid adaptation to new MRI modalities (e.g., T2w, T1c) or to other lesion types such as traumatic brain injury and multiple sclerosis.
The codebase is organised into three repositories: (i) the C++ bootstrapper, (ii) the build and packaging scripts using CMake/CPack, and (iii) the core Python scientific package. The CMake/CPack repository provides a simple, dependency-free build process that automatically bundles all required components. This modular structure facilitates integration into broader clinical or research pipelines, ensuring transparency, portability, and maintainability while providing a lightweight alternative to Docker-based deployment.
The tool runs on Windows and Linux, with macOS support planned pending further CPack development. Key features include automatic viewer launch for the first processed subject, probability map export, configurable lesion thresholds, and optional MNI-space outputs. System paths for logs, models, and configuration follow OS conventions: \texttt{AppData} and \texttt{ProgramData} on Windows; \texttt{$\sim$/.local/share} and \texttt{$\sim$/.config} on Linux.

\section{Results}
\subsection{Accuracy}
\label{sec:accuracy}
Quantitative validation on 300 unseen stroke subjects (ATLAS v2.0~\cite{liew2018atlas} test-set) verified numerical equivalence between PyTorch nnU-Net and the \texttt{Float16} ONNX implementation. A mean Dice difference of $<10^{-3}$ (lowest agreement $0.9977$) confirmed model behavior preservation. Lesion Dice scores were consistent with published results~\cite{dimatteo2025stroke}, with mean values $68.8$ (T1w) and $75.6$ (T1w+FLAIR).
\subsection{Performance and size}
\label{sec:size}
Packaged application sizes are reported in Table~\ref{tab:size}. ONNX conversion reduced overall package size by approximately 60\% compared with the original PyTorch builds.
\vspace{-0.5cm}
\begin{table}[ht]
\centering
\caption{Packaged application size comparison.}
\label{tab:size}
\begin{tabular}{lccc}
\toprule
\textbf{OS (Hardware)} & \textbf{PyTorch} & \textbf{ONNX} & \textbf{Reduction} \\
\midrule
Fedora (GPU+CPU)   & 6.20 GB & 2.50 GB & $\sim$60\% \\
Windows (GPU+CPU)  & 7.25 GB & 2.73 GB & $\sim$62\% \\
Windows (CPU only)  & 2.10 GB & 0.80 GB & $\sim$62\% \\

\bottomrule
\end{tabular}
\end{table}
\vspace{-1cm}
\begin{table}[ht]
\centering
\caption{Runtime comparison across processing stages.}
\label{tab:runtime}
\begin{tabular}{lcc}
\toprule
\textbf{Task} & \textbf{Ultrabook laptop} & \textbf{Workstation laptop} \\
\midrule
  Brain Extraction & $198.5$~s/volume & $36.6$~s/volume \\
  Registration & $15.1$~s/volume & $3.1$~s/volume \\
  Inference & $149.2$~s/volume & $0.9$~s/volume \\

\bottomrule
\end{tabular}
\end{table}

Execution times are summarised in Table~\ref{tab:runtime}. Averaged end-to-end processing time per subject was 6.0~min on the CPU-only laptop and 41~s on the GPU workstation. As expected, performance was strongly affected by hardware configuration. On the workstation, GPU-accelerated inference achieved a $\sim$165$\times$ speedup over the CPU-only laptop ($0.9$~s vs.~$149.2$~s per volume). Preprocessing steps showed more moderate but still substantial improvements: brain extraction was 5.4$\times$ faster (36.6~s vs.~198.5~s per volume), and registration improved by 4.9$\times$ (3.1~s vs.~15.1~s per volume). These results confirm that inference is the most GPU-sensitive stage, while brain extraction - limited by CPU performance per core~\cite{Amdahl1967} - remains the dominant contributor to total execution time. Caching or importing preprocessed volumes therefore provides the most effective way to reduce overall processing time in repeated analyses.

\section{Discussion and Conclusion}
StrokeSeg demonstrates that research-grade nnU-Net models can be transformed into a practical, lightweight tool without compromising accuracy. This confirms that engineering optimisations can achieve deployability gains while preserving scientific fidelity. No systematic performance differences were observed across lesion sizes, indicating that precision quantisation did not preferentially affect small or large lesions. The use of ONNX Runtime and \texttt{Float16} quantisation substantially reduces software dependencies and package size, while the modular design and dual GUI/CLI interfaces improve usability across research and clinical contexts.

\textbf{Limitations.} Current constraints are (i) lack of DICOM I/O  (NIfTI only), (ii) restricted throughput due to serial execution, and (iii) research-only without CE/FDA clearance.

\textbf{Future work.} 
Upcoming developments will address limitations (i) and (ii), and introduce a native C++/Qt interface for improved usability. We also plan a full C++ implementation using ONNX Runtime to further reduce footprint and improve portability. These efforts aim toward a coherent, lightweight, standalone C++ application. We intend to support heterogeneous inference backends across diverse hardware (AMD, NVIDIA, Intel), including integrated GPUs commonly used in clinical environments and NPU built-in CPUs.

\section{Compliance with ethical standards}
This study used retrospectively processed MRI data from the publicly available ATLAS v2.0 dataset~\cite{liew2018atlas}, together with pretrained models~\cite{dimatteo2025stroke}. In accordance with the licensing terms of the dataset, additional ethical approval is not required.

\vspace{4pt}
\textbf{Software availability}:
StrokeSeg executables and guide are provided as research software at~\cite{StrokeSegUserGuide2025}. The application displays a startup disclaimer and is not CE-marked.

\textbf{Conflicts of Interest}: The authors have no relevant financial or non-financial interests to disclose.

\bibliographystyle{IEEEbib}
\bibliography{strings,refs}

\end{document}